\documentclass{pbml}

\usepackage{float}

\usepackage[table]{xcolor}
\usepackage{comment}
\usepackage{todonotes}

\begin{document}


\title{Fine-grained human evaluation of neural \titlelinebreak{}versus phrase-based machine translation}


\institute{label1}{University of Zagreb}
\institute{label2}{University of Groningen}
\institute{label3}{Prompsit Language Engineering}

\author{
  firstname=Filip,
  surname=Klubi\v{c}ka,
  corresponding=yes,
  institute=label1,
  email={fklubick@ffzg.hr},
  address={Faculty of Humanities and Social Sciences\\3 Ivan Lu\v{c}i\'{c} Street, Zagreb, PA 10000, Republic of Croatia}
}

\author{
  firstname=Antonio,
  surname=Toral,
  institute={label2},
}

\author{
  firstname=Víctor,
  initials=M.,
  surname=Sánchez-Cartagena,
  institute=label3,
}


\shorttitle{Human evaluation of NMT vs. PBMT}
\shortauthor{F. Klubi\v{c}ka, A. Toral, V. M. Sánchez-Cartagena}

\PBMLmaketitle


\begin{abstract}

We compare three approaches to statistical machine translation (pure phrase-based, factored phrase-based and neural) by performing a fine-grained manual evaluation via error annotation of the systems' outputs. The error types in our annotation are compliant with the multidimensional quality metrics (MQM), and the annotation is performed by two annotators. Inter-annotator agreement is high for such a task, and results show that the best performing system (neural) reduces the errors produced by the worst system (phrase-based) by 54\%.

\end{abstract}

\section{Introduction}\label{s:intro}

A paradigm to machine translation (MT) based on deep neural networks and usually referred to as neural MT (NMT)
has emerged in the past few years.
This has disrupted the MT field as NMT, despite its infancy, has already surpassed the performance of phrase-based MT (PBMT), the mainstream approach to date.

We have witnessed the potential of NMT in terms of overall performance scores, be those automatic (e.g. BLEU) or human (e.g. system rankings); for example, in last year's news translation shared task at WMT.\footnote{\url{http://www.statmt.org/wmt16/translation-task.html}}
There, out of 9 language directions where NMT systems were submitted, they significantly outperformed PBMT in 8, according to the human evaluation.
In the remaining language direction (Russian-to-English), the best PBMT submission was ranked higher than the best NMT system, but the difference was found not to be significant.

Given the impressive overall performance of NMT, some researchers have attempted in the past year to analyse the potential of NMT in a more detailed manner.
The motivation comes from the fact that
while overall scores give an indication of the general performance of a system, they do not provide any additional information.
Hence, in order to delve further and try to shed light on the strengths and weaknesses of this new paradigm to MT, two recent papers have looked at conducting multifaceted evaluations.
\begin{itemize}
\item 
\citet{D16-1025} conducted a detailed analysis for the English-to-German language direction where they compared state-of-the-art PBMT and NMT systems on transcribed speeches.
They found out that NMT (i) decreases post-editing effort, (ii) degrades faster than PBMT with sentence length and (iii) improves notably on reordering and inflection.

\item
\citet{DBLP:journals/corr/ToralS17} carried out a series of analyses and evaluations for NMT and PBMT systems on the domain of news for 9 language pairs.
They corroborated the findings of \citet{D16-1025} with respect to NMT outstanding performance on reordering and inflection and its degradation with sentence length.
They also contributed additional findings: NMT systems (i) exhibit higher inter-system variability, (ii) lead to more fluent outputs and (iii) perform more reordering than PBMT but less than hierarchical PBMT.
\end{itemize}

A limitation of these analyses lies in the fact that all of them were performed automatically. E.g. reordering and inflection errors were detected based on automatic evaluation metrics.
Hence, one could argue that their outcomes are somewhat affected as automatic tools are, of course, never perfect.

In this paper we conduct a detailed human analysis of the outputs produced by NMT and PBMT systems.
Namely, we annotate manually the errors found according to a detailed error taxonomy, that is compliant with the hierarchical listing of issue types defined as part of the Multidimensional Quality Metrics (MQM)~\cite{mqm}. Specifically, we carry out this analysis for the news domain in the English-to-Croatian language direction.
First, we define an error taxonomy that is relevant to the problematic linguistic phenomena of this language pair.
Subsequently, we annotate the errors produced by 3 state-of-the-art translation systems that belong to the following paradigms: PBMT, factored PBMT and NMT.
Finally, we analyse the annotations.

The main contributions of this paper can then be summarised as follows:
\begin{enumerate}
\item We conduct, to the best of our knowledge, the first human fine-grained error analysis of NMT in the literature.
\item We analyse NMT in comparison not only to pure PBMT and hierarchical PBMT, as in previous works, but also with respect to factored models.
\item We develop an MQM-compliant error taxonomy for Slavic languages.
\item We develop a novel approach to statistically analyzing and interpreting the results of MQM error annotation.

\end{enumerate}

The rest of the paper is organised as follows.
Section \ref{s:sys} describes the MT systems and the datasets used in our experiments.
Section \ref{s:mqm} covers the analysis, including the definition of the error taxonomy, the annotation setup and guidelines and finally the results obtained and their discussion.
Finally, Section \ref{s:con} outlines the conclusions and lines of future work.

\section{MT Systems}\label{s:sys}

This section describes the MT systems and the datasets used in our experiments. We built PBMT, factored PBMT and NMT systems.

The 3 systems were trained on the same parallel data.
We considered a set of publicly available English--Croatian parallel corpora, comprising the DGT Translation Memory~\footnote{\url{https://ec.europa.eu/jrc/en/language-technologies/dgt-translation-memory}}, HrEnWaC\footnote{\url{https://www.clarin.si/repository/xmlui/handle/11356/1058}}, JRC Acquis~\footnote{\url{http://tinyurl.com/CroatianAcquis}}, OpenSubtitles 2013, \textsc{SETimes} and \textsc{Ted} talks.
We concatenated all these corpora and performed cross-entropy based data selection~\citep{Moore:2010:ISL:1858842.1858883} using the development set.
Once the data is ranked we keep the highest ranked 25\% sentence pairs (4,786,516).

PBMT systems used also monolingual data for language modelling.
To this end we used the concatenation of the hrWaC corpus \cite{ljubesic14-bs} 
and the target side of the aforementioned parallel corpora.

As development set we used the first 1,000 sentences of the English test set used at the WMT12 news translation task\footnote{\url{http://www.statmt.org/wmt12/translation-task.html}}, translated by a professional translator into Croatian.
Similarly, our test set is made of the first 1,000  sentences of the English test set of the WMT13 translation task\footnote{\url{http://www.statmt.org/wmt13/translation-task.html}}, again manually translated into Croatian.

The PBMT system was built with Moses v3.0\footnote{\url{https://github.com/moses-smt/mosesdecoder/tree/RELEASE-3.0}}.
In addition to the default models we also used
hierarchical reordering~\citep{galley2008simple}, an operation sequence model~\citep{durrani2011joint} and a bilingual neural language model~\citep{devlin2014binlm}.

The factored PBMT system maps one factor in the source language (surface form) to two factors in the target (surface form and morphosyntactic description).
This system is described in detail by~\citet{W16-3421}.

The NMT system is based on the sequence-to-sequence architecture with attention and we applied sub-word segmentation with byte pair encoding~\citep{sennrich2015a} jointly on the source and target languages.
We performed $85\,000$ join operations.
Training was run for $10$ days and a model was saved every $4.5$ hours.
We decoded the test set using an ensemble of 4 models. These were the 4 models with the highest BLEU scores on the development set.

\subsection{Evaluation}

We report the scores obtained in terms of the BLEU and TER automatic evaluation metrics for the 3 systems described in the previous section.
Table \ref{t:autoeval} shows the results.

As the table shows, the use of factored models leads to a substantial improvement upon pure PBMT (6\% relative in terms of BLEU).
NMT, on its turn, allows us to obtain a further notable improvement; 14\% relative in terms of BLEU compared to the factored PBMT system and 21\% compared to the initial PBMT system.

\begin{table}[htbp]
\begin{center}
\begin{tabular}{lrr}
\hline
\bf System & \bf BLEU & \bf TER\\
\hline 
PBMT & 0.2544 & 0.6081\\
Factored PBMT &0.2700 &0.5963\\
NMT &0.3085 & 0.5552\\
\hline 
\end{tabular}
\caption{Automatic evaluation (BLEU and TER scores) of the 3 MT systems
\label{t:autoeval}}
\end{center}
\end{table}

\section{Error analysis}\label{s:mqm}

In this section we report on the motivation for conducting the manual error analysis, describe the framework and overall annotation process, and present the results.

The fact that Croatian is rich in inflection, has rather free word order and other similar phenomena that English does not, gives rise to specific translation issues. For example, grammatical categories that do not exist in English, like gender and case, may be particularly hard to generate reliably in a Croatian translation. We built our factored PBMT system aiming to directly address such issues. Similarly motivated, we wished to see how an NMT system would grapple with the same issues.

Indeed, as shown in Section \ref{s:sys}, automatic evaluation shows significant improvement for both systems, compared to the pure PBMT system. However, as is the nature of automatic metrics, the automatic scoring methods do not indicate whether any of the linguistic problems mentioned earlier have been addressed by the systems. The question of whether the linguistic quality, or rather, grammaticality of the output is improved has not been answered by automatic evaluation. Are cases and gender handled better? Is there better agreement? Is the fluency of the translation higher?

In order to provide answers to these research questions, we decide to thoroughly compare these systems by systematically analyzing their outputs via manual error analysis. In this way we can obtain a more complete picture of what is happening in the translation, which can provide pointers on where to act to obtain further improvements in the future.

\subsection{Multidimensional Quality Metrics and the Slavic tagset}

After looking into different ways of performing the task of manual evaluation via error analysis, we decided to make use of the MQM framework, developed in the QTLaunchpad project\footnote{\url{http://www.qt21.eu/mqm-definition/definition-2015-06-16.html}}. This is a framework for describing and defining custom translation quality metrics. It provides a flexible vocabulary of quality issue types and a mechanism for applying them to generate quality scores. It does not impose a single metric for all uses, but rather provides a comprehensive catalog of quality issue types, with standardized names and definitions, that can be used to describe particular metrics for specific tasks.

The main reason we chose the MQM framework was the flexibility of the issue types and their granularity --- it gave us a reliable methodology for quality assessment, that still allowed us to pick and choose which error tags we wish to use. 

\begin{figure}[h!]
	\centering
	\includegraphics[width=0.9\textwidth]{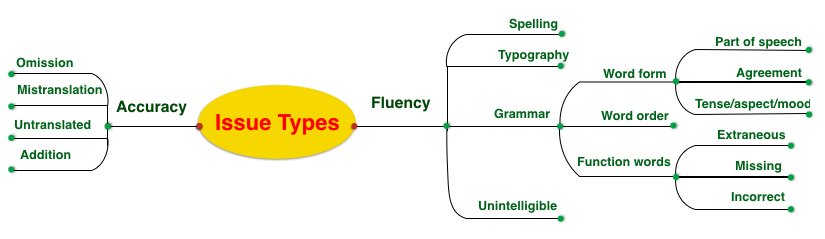}
	\caption{The core error categories proposed by the MQM guidelines
    \label{fig:core_set}}
\end{figure}

The MQM guidelines propose a great variety of tags on several annotation layers\footnote{\url{http://www.qt21.eu/mqm-definition/issues-list-2015-12-30.html}}. However, the full tagset is too comprehensive to be viable for any annotation task, so the process begins with choosing the tags to use in accordance to our research questions. Initially we started off with the core tagset, a default set of evaluation metrics (i.e.\ error categories) proposed by the MQM guidelines, as seen in Figure \ref{fig:core_set}.

\begin{figure}[h!]
	\centering
	\includegraphics[width=0.7\textwidth]{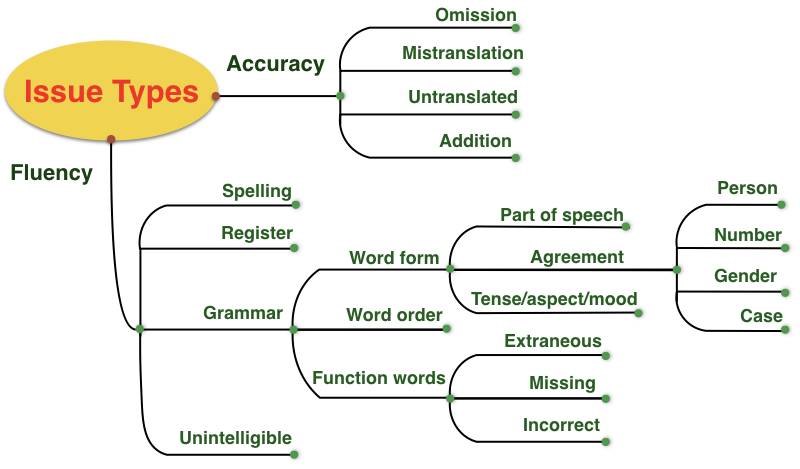}
	\caption{The Slavic tagset, a modified version of the MQM core tagset
    \label{fig:slavic_set}}
\end{figure}

However, given the morphological complexity of Croatian and the level at which we made interventions in the system, we found that these core categories were not detailed enough, or rather, did not allow for an analysis of the specific phenomena we were interested in. Some categories that were of interest to us, like specific \textit{Agreement} types, were not present in the tagset, while some errors, like \textit{Typography}, were irrelevant to us.
So we created our own set of tags by modifying the core set, rearranging the hierarchy, adding new tags and removing those that are of little relevance.
We call this new tagset the Slavic tagset, as its expansion allows for the identification of grammatical errors which are commonly shared by Slavic languages. This tagset is outlined in Figure \ref{fig:slavic_set}.

\subsection{Annotation setup}

In order to carry out the annotations we used \texttt{translate5}\footnote{\url{http://www.translate5.net/}}, a web-based tool that implements annotations of MT outputs using hierarchical taxonomies, as is the case of MQM.

We had two annotators at our disposal, who both had prior experience with MQM as well as the same background - an MA in English linguistics and information science. They were thoroughly familiarized with the official annotation guidelines and the decision process\footnote{\url{http://www.qt21.eu/downloads/annotatorsGuidelines-2014-06-11.pdf}} prior to annotation.

The annotators annotated 100 random sentences from the test set introduced in Section \ref{s:sys}. These sentences were translated by all three MT systems, and the annotators were presented with the source text, a reference translation and the unannotated system outputs at the same time. All three translations were then annotated by both our annotators (i.e. each system translated the same 100 sentences, each annotator annotated the 300 translated sentences, making a total of 600 annotated sentences).
Once the sentences were annotated, the annotation data was extracted, we calculated inter-annotator agreement and analyzed the output to see what the number of error tags can tell us about the performance of each system.

\subsection{Inter-Annotator Agreement}

Though carefully thought out and developed, the MQM metrics, and manual MT evaluation in general, are notorious for resulting in low inter-annotator agreement scores. This is attested by the body of work that has addressed this issue, most notably \citet{lommel-iaa}, who worked specifically on MQM, and \cite{callison-iaa}, who investigated several tasks. This is why it is important that we check how well our annotators agree on the task at hand, and whether this is consistent with other work done with MQM so far.

Once the data was annotated, agreement was observed at the sentence level, and inter-annotator agreement was calculated using the Cohen's Kappa ($\kappa$) metric \citep{kappa}.
Agreement was calculated on the annotations of every system separately, as well as on a concatenation of annotations, in order to both see whether there are differences in agreement across systems, as well as to gain insight into the overall agreement between annotators. Additionally, Coehn's $\kappa$ was also calculated for every error type separately. Detailed results can be found in Table \ref{t:IAA}.

\begin{table}[htbp]
\begin{center}
\begin{tabular}{lrrrr}
\hline
\bf Error type & \bf PBMT & \bf Factored & \bf NMT & \bf Concatenated\\
\hline
Accuracy \\
\hspace*{1em}Mistranslation & 0.51 & 0.48 & {\bf 0.58} & 0.53 \\
\hspace*{1em}Omission & 0.34 & 0.39 & 0.37 & 0.37 \\
\hspace*{1em}Addition & 0.5 & 0.54 & 0.33 & 0.47 \\
\hspace*{1em}Untranslated & {\bf 0.86} & {\bf 0.86} & -0.02 & \bf 0.72 \\
Fluency \\
\hspace*{1em}Unintelligible & 0.39 & 0.32 & 0 & 0.35 \\
\hspace*{1em}Register & 0.37 & 0.2 & 0.22 & 0.27 \\
\hspace*{1em}Word order & 0.56 & 0.33 & 0.21 & 0.4 \\
\hspace*{1em}Function words \\
\hspace*{2em}Extraneous & 0.56 & 0.32 & 0.49 & 0.46 \\
\hspace*{2em}Incorrect & 0.37 & 0.18 & 0.34 & 0.29 \\
\hspace*{2em}Missing & 0 & 0.49 & 0 & 0.33 \\
\hspace*{1em}Tense... & 0.44 & 0.36 & 0.15 & 0.38 \\
\hspace*{1em}Agreement & 0.24 & 0.41 & 0 & 0.33 \\
\hspace*{2em}Number & 0.53 & 0.55 & 0.52 & 0.54 \\
\hspace*{2em}Gender & 0.46 & 0.59 & 0.48 & 0.53 \\
\hspace*{2em}Case & 0.53 & 0.49 & 0.52 & 0.56 \\
\hline
\bf All errors & \bf 0.56 & \bf 0.49 & \bf 0.44 & \bf 0.51 \\
\hline 
\end{tabular}
\caption{Inter-annotator agreement (Cohen's $\kappa$ values) for the MQM evaluation task. The highest score for any individual system and the concatenation, as well as the overall score, are shown in bold.
\label{t:IAA}}
\end{center}
\end{table}

Generally, one can see that our annotators agree best on evaluations of the PBMT system, less so on evaluations of the Factored SMT system, and least in evaluations of the NMT system. Overall agreement scores are relatively low - the average total $\kappa$ is approximately 0.51. Furthermore, the $\kappa$ scores are relatively consistent across all error types, mostly ranging between 0.35 and 0.55. 
According to Cohen, such scores constitute moderate agreement.
However, as already stated, this is to be expected, given the complexity of the problem and annotation schema. In fact, this is a notably higher score than what has been reported in similar work, e.g. \citet{lommel-iaa}, who achieve 
$\kappa$ scores ranging between 0.25 and 0.34.
However, this comparison should be taken with a grain of salt, as our calculations are just an approximation compared to Lommel et al.'s, given that in our setup we looked only at sentence level agreement, while they calculated agreement on the token level.  


\subsection{Results of annotation
\label{sub:results}}

Directly extracting raw annotation data from the \texttt{translate5} system provides a sum of error tags annotated for each error type by each annotator and system. The total values are presented in Table \ref{t:MQM-raw}. 

\begin{table}[htbp]
\begin{center}
\begin{tabular}{lrrr|rrr}
\hline
  & \multicolumn{3}{c|}{\bf Annotator 1}  & \multicolumn{3}{c}{\bf Annotator 2} \\
\hline
System  & \bf PBMT & \bf Factored & \bf NMT & \bf PBMT & \bf Factored & \bf NMT \\
\hline
Total errors & 317 & 276 & 178 & 264 & 199 & 132 \\
\hline
\end{tabular}
\caption{Total errors per system per annotator
\label{t:MQM-raw}}
\end{center}
\end{table}

Looking at the aggregate data alone, one can easily detect that both annotators have judged that the PBMT system contains the most errors, and that the NMT system contains the smallest number of errors. This trend is consistent across most fine-grained error categories as well.

However, even though simply counting the errors can provide insight into which system performs better, we thought that this approach does not adequately represent our findings, as it does not allow a proper quantification of the quality of the outputs. Certainly, based on data from Table \ref{t:MQM-raw} we can claim, for example, that the NMT system produces less errors in general, or less errors of a specific type, but given that the outputs are different, as is the number of tokens in each translation, 
we decided to normalize the data.

To the best of our knowledge there is no related work on how to approach this, as previous work simply counts the number of MQM tags and stops there. After some consideration, we decided to normalize at the token level.
I.e. instead of counting just error tags produced by each annotator, we count the tokens that these errors are assigned to -- tokens that do and tokens that do not have an error annotation.
Once these numbers are divided by the total number of tokens in the system's output, they provide a concrete idea of the ratio of tokens with and without errors.

The results of such analysis again show that the PBMT system has the largest error ratio, while the NMT system has the smallest one.
This is further backed up by a pairwise chi-squared ($\chi^2$) statistical significance test; we calculate statistical significance from 2x2 contingency tables for every system pair (PBMTxFactored, PBMTxNMT and FactoredxNMT). The results show that the differences in the total number of tokens with errors are statistically significant for all three system pairs, with the $p$ value being lower than 0.0001 in each case.

Furthermore, we also wanted to see which error types are the ones making a significant impact on this result.
So we repeated these same measurements, but instead of performing them on all error types combined, they were performed separately for each specific error category.
The combined results of the calculations and transformations are presented in Table \ref{t:MQM-token}.

\begin{table}[htbp]
\begin{center}
\begin{tabular}{lrr|rr|rr}
\hline
  & \multicolumn{2}{c|}{\bf PBMT}  & \multicolumn{2}{c|}{\bf Factored} & \multicolumn{2}{c}{\bf NMT} \\ 
\hline
\bf Error type & \bf No error & \bf Error & \bf No error & \bf Error & \bf No error & \bf Error \\
\hline
Accuracy & 3467 & 369 & \cellcolor{green!25}3525 & \cellcolor{green!25}*291 & 3402 & 266 \\
\hspace*{1em}Mistranslation & 3547 & 289 & \cellcolor{green!25}3586 & \cellcolor{green!25}*230 & 3471 & 197 \\
\hspace*{1em}Omission & 3801 & 35 & 3793 & 23 & \cellcolor{red!25}3619 & \cellcolor{red!25}*49 \\
\hspace*{1em}Addition & 3814 & 22 & 3797 & 19 & 3655 & 13 \\
\hspace*{1em}Untranslated & 3813 & 23 & 3797 & 19 & \cellcolor{green!25}3662 & \cellcolor{green!25}*6 \\
  &  &  &  &  &  & \\
Fluency & 3195 & 641 & \cellcolor{green!25}3298 & \cellcolor{green!25}*518 & \cellcolor{green!25}3465 & \cellcolor{green!25}**188 \\
\hspace*{1em}Unintelligible & 3790 & 46 & 3769 & 47 & \cellcolor{green!25}3668 & \cellcolor{green!25}**0 \\
\hspace*{1em}Register & 3810 & 26 & 3794 & 22 & 3646 & 22 \\
\hspace*{1em}Spelling & 3833 & 3 & 3812 & 4 & 3659 & 9 \\
\hspace*{1em}Grammar & 3270 & 566 & \cellcolor{green!25}3371 & \cellcolor{green!25}**445 & \cellcolor{green!25}3497 & \cellcolor{green!25}**156 \\
\hspace*{1em}Word order & 3752 & 84 & 3752 & 64 & \cellcolor{green!25}3646 & \cellcolor{green!25}**22 \\
\hspace*{1em}Function words & 3801 & 35 & 3780 & 36 & \cellcolor{green!25}3650 & \cellcolor{green!25}*18 \\
\hspace*{2em}Extraneous & 3829 & 7 & 3810 & 6 & 3664 & 4 \\
\hspace*{2em}Incorrect & 3810 & 26 & 3790 & 26 & \cellcolor{green!25}3655 & \cellcolor{green!25}*13 \\
\hspace*{2em}Missing & 3834 & 2 & 3812 & 4 & 3667 & 1 \\
\hspace*{1em}Word form & 3389 & 447 & \cellcolor{green!25}3471 & \cellcolor{green!25}*345 & \cellcolor{green!25}3538 & \cellcolor{green!25}**102 \\
\hspace*{1em}Part of speech & 3822 & 14 & 3800 & 16 & \cellcolor{green!25}3663 & \cellcolor{green!25}*5 \\
\hspace*{1em}Tense... & 3775 & 61 & 3765 & 51 & \cellcolor{green!25}3648 & \cellcolor{green!25}*20 \\
\hspace*{1em}Agreement & 3466 & 370 & \cellcolor{green!25}3540 & \cellcolor{green!25}*276 & \cellcolor{green!25}3566 & \cellcolor{green!25}**102 \\
\hspace*{2em}Number & 3778 & 58 & 3772 & 44 & \cellcolor{green!25}3646 & \cellcolor{green!25}*22 \\
\hspace*{2em}Gender & 3788 & 48 & 3756 & 60 & \cellcolor{green!25}3644 & \cellcolor{green!25}*24 \\
\hspace*{2em}Case & 3614 & 222 & \cellcolor{green!25}3694 & \cellcolor{green!25}*122 & \cellcolor{green!25}3622 & \cellcolor{green!25}**46 \\
\hspace*{2em}Person & 3836 & 0 & 3816 & 0 & 3664 & 4 \\
\hline
Total errors & 2826 & 1010 & \cellcolor{green!25}3007 & \cellcolor{green!25}**809 & \cellcolor{green!25}3199 & \cellcolor{green!25}**469 \\
\hline 
\end{tabular}
\caption{Processed annotation data from both annotators concatenated: each system's total number of tokens with and without errors. Statistical significance for a system, when compared to the system on its left, is marked with * where $p$-value is \textless0.05 and ** where $p$-value is \textless0.0001. Cells with a green background indicate that the system has less errors than the one on its left, while those in red indicate that it has more.
\label{t:MQM-token}}
\end{center}
\end{table}

We can derive several findings from this table.  
Firstly, when looking simply at the grand total of tokens with and without errors, the difference between the systems is statistically significant by a wide margin. 
When looking at PBMT and factored PBMT, the factored system has significantly less errors than the pure PBMT system. The overall error rate is in this case reduced by 20\%.
A separate analysis of specific error types that contribute to this score reveals that only some of the error categories are significantly different between the two systems. In the table, those categories are filled in with green. One can see that, when it comes to agreement, the only agreement type that produces significantly less errors is agreement in case.

However, taking a look at NMT shows that, not only does it result in a 42\% overall error reduction compared to the factored system, and 54\% with respect to pure PBMT, but it produces even less agreement errors -- overall, as well as at the level of number, gender and case -- while not using any kind of linguistic information at all. This might in part be due to the use of sub-word segmentation, as inflections in Croatian are relatively regular. 
In addition to improving in the Agreement category, NMT also produces significantly less errors in many more categories than the factored model does.
Interestingly, it produces more Omission errors than either of the other two systems. It seems that it tends to sacrifice completeness of translation in order to increase overall fluency. Indeed, extrapolating from the data in Table \ref{t:MQM-token}, shows that, though differences are very small, NMT does have the lowest token per sentence ratio (PBMT 18.99, Factored PBMT 18.89, NMT 18.36). 

\section{Conclusion}\label{s:con}

The fine-grained manual evaluation performed for the purpose of this research has provided answers to several questions, one of which was the main drive behind our developing the factored system: is there a way to handle better agreement when translating to Croatian? We can now confidently claim that factored models result in significantly less agreement errors overall compared to pure PBMT.

We can also confidently claim that NMT handles all types of agreement better than both pure PBMT and factored PBMT, which corroborates the findings of other researchers' NMT evaluations. Our system produces sentences with far less errors, and a language that is more fluent and more grammatical, which should be of help when it comes to the task of post-editing. 

Furthermore, the error taxonomy that was developed for this research, while only used for the English-to-Croatian language direction, should be applicable for the analysis of errors for any translation direction towards a Slavic language, as it takes into account grammatical properties specific to these languages.

Among other possible lines of future work, including the application of our methodology to another language pair (e.g. English-Czech), performing more controlled IAA analysis or IAA adjudication, as well as comparing to an NMT model without sub-word segmentation, another one is adapting the tagset further. In its current version, it has proved to be informative when comparing PBMT to factored PBMT. However, NMT has shown itself to produce language that is so fluent that the fine-grained hierarchy in the \textit{Fluency} branch is of little use. Meanwhile, the most common error type in the NMT output is \textit{Mistranslation}, which, according to the MQM guidelines, covers both lexical selection and, less intuitively, translation of grammatical properties (e.g. if 'cats[pl.]' is translated as 'mačka[sg.]', this is to be tagged as \textit{Mistranslation}, in spite of correct lexical choice). This makes it quite a vague category, so if one would wish to perform an even more nuanced linguistic error analysis for NMT, adding additional layers to the \textit{Accuracy} branch would seem a promising direction to follow.

\section*{Acknowledgements}
We would like to extend our thanks to Maja Popović, who provided valuable advice on how to approach the annotation and evaluation, and Denis Kranjčić, who participated in the annotation task. The research leading to these results has received funding from the European Union Seventh Framework Programme FP7/2007-2013 under grant agreement PIAP-GA-2012-324414 (Abu-MaTran) and the Swiss National Science Foundation grant 74Z0\_160501 (ReLDI).

\bibliography{mybib}


\end{document}